%% file: main.tex
\pdfoutput=1

\documentclass[11pt]{article}

\usepackage[final]{acl}

\usepackage{times}
\usepackage{latexsym}
\usepackage{url}
\usepackage{multirow}
\usepackage{enumitem}
\usepackage{adjustbox}
\usepackage{tcolorbox}
\usepackage{booktabs} 
\usepackage{amsfonts}
\usepackage[T1]{fontenc}
\usepackage[utf8]{inputenc}
\usepackage{microtype}
\usepackage{inconsolata}
\usepackage{graphicx}
\usepackage{xspace}

\setlist{leftmargin=*,nosep}

\title{NyayaRAG: Realistic Legal Judgment Prediction with RAG under the Indian Common Law System}

\author{Shubham Kumar Nigam$^{1,5*\dagger}$ \hskip2mm Balaramamahanthi Deepak Patnaik$^{1*}$ \hskip2mm Shivam Mishra$^{1*}$\\ 
\textbf{Ajay Varghese Thomas}$^{2*}$ \hskip2mm \textbf{Noel Shallum}$^{3}$ \hskip2mm \textbf{Kripabandhu Ghosh}$^{4}$ \hskip2mm \textbf{Arnab Bhattacharya}$^{1}$\\
$^{1}$Indian Institute of Technology Kanpur, India \quad
$^{2}$SRM Institute of Science and Technology, India \\
$^{3}$Symbiosis Law School Pune, India \quad
$^{4}$IISER Kolkata, India \\
$^{5}$University of Birmingham, Dubai, United Arab Emirates\\
\texttt{\{shubhamkumarnigam, bdeepakpatnaik2002, shivam1602m,}\\
\texttt{ajaythomas.work, noelshallum\}@gmail.com} \\ 
\quad \texttt{kripaghosh@iiserkol.ac.in} \quad \texttt{arnabb@cse.iitk.ac.in}
}

\date{}

\begin{document}

\maketitle

{
\renewcommand{\thefootnote}{$*$}
\footnotetext{These authors contributed equally to this work}
\renewcommand{\thefootnote}{$\dagger$}
\footnotetext{Corresponding author}
\renewcommand{\thefootnote}{\arabic{footnote}}
}

\input{abstract}

\input{intro}

\input{related_work}

\input{task_description}

\input{dataset}

\input{methodology}

\input{evaluation_metrics}

\input{results_and_analysis}

\input{Ablation_study}

\input{conculsion}

\input{acknowledgement}

\input{limitation}

\input{ethics}
\newpage
\bibliography{anthology, custom, sknigam, legal_bib}

\newpage
\appendix
\input{appendix}

\end{document}

%% file: abstract.tex
\begin{abstract}
Legal Judgment Prediction (LJP) has emerged as a key area in AI for law, aiming to automate judicial outcome forecasting and enhance interpretability in legal reasoning. While previous approaches in the Indian context have relied on internal case content such as facts, issues, and reasoning, they often overlook a core element of common law systems -- reliance on statutory provisions and judicial precedents. In this work, we propose \textbf{\texttt{NyayaRAG}}, a Retrieval-Augmented Generation (RAG) framework that simulates realistic courtroom scenarios by providing models with factual case descriptions, relevant legal statutes, and semantically retrieved prior cases. NyayaRAG evaluates the effectiveness of these combined inputs in predicting court decisions and generating legal explanations using a domain-specific pipeline tailored for the Indian legal system. We assess performance across various input configurations using both standard lexical and semantic metrics as well as LLM-based evaluators such as G-Eval. Our results show that augmenting factual inputs with structured legal knowledge significantly improves both predictive accuracy and explanation quality.
\end{abstract}

%% file: intro.tex
\section{Introduction}
The application of artificial intelligence (AI) in legal judgment prediction (LJP) has the potential to transform legal systems by improving efficiency, transparency, and access to justice. This is particularly crucial for India, where millions of cases remain pending in courts, and decision-making is inherently dependent on factual narratives, statutory interpretation, and judicial precedents. India follows a common law system, where prior decisions (precedents) and statutory provisions play a central role in influencing legal outcomes. However, most existing AI-based LJP systems do not adequately replicate this fundamental feature of judicial reasoning.

Previous studies \citep{malik-etal-2021-ildc, nigam-etal-2024-legal, nigam-etal-2025-nyayaanumana} have focused on predicting legal outcomes using the current case document, including sections like facts, arguments, issues, reasoning, and decision. More recent efforts have narrowed the scope to factual inputs alone \citep{nigam-etal-2024-rethinking, FactLegalLlama}; yet these systems still operate in a setting that is not real, and does not consider how courts naturally rely on applicable laws and prior rulings. In reality, judges rarely decide in isolation; instead, they actively refer to relevant precedents and statutory laws. To bridge this gap, we propose a framework that more closely mirrors actual courtroom conditions by explicitly incorporating external legal knowledge during inference.

In critical domains like law, medicine and finance, decisions must be grounded in verifiable information. Experts in these domains cannot rely on opaque, black-box inferences and require systems that ensure factual consistency. Hallucinations, common in large generative models, can have severe consequences in legal decision-making. By retrieving and conditioning model responses on grounded sources such as applicable laws and precedent cases, the Retrieval-Augmented Generation (RAG) paradigm offers a principled approach to mitigate hallucination and promote trustworthy outputs. Furthermore, RAG frameworks like ours can be flexibly integrated into existing legal systems without requiring the retraining of core models or the sharing of private or sensitive case data. This enhances user trust while allowing the legal community to benefit from AI without sacrificing transparency or data confidentiality.

Motivated by the above, we introduce \textbf{\texttt{NyayaRAG}}, a Retrieval-Augmented Generation (RAG) framework for realistic legal judgment prediction and explanation in the Indian common law system. The term ``NyayaRAG'' is derived from two components: ``\textbf{Nyaya}'' meaning ``justice'' and ``\textbf{RAG}'' referring to ``Retrieval-Augmented Generation''. Together, the name reflects our vision to build a justice-aware generation system that emulates the reasoning process followed by Indian courts, using facts, statutes, and precedents.

Unlike prior models that operate purely on internal case content, \texttt{NyayaRAG} simulates real-world judicial decision-making by providing the model with:
(i)~the \emph{summarized factual background} of the current case,
(ii)~\emph{relevant statutory provisions}, and (iii)~retrieved top-k \emph{semantically similar previous judgments}. This structure emulates how judges deliberate on new cases, consulting both textual statutes and prior judicial opinions. Through this design, we evaluate how RAG can help reduce hallucinations, promote faithfulness, and yield legally coherent predictions and explanations.

Our contributions in this paper are as follows:
\begin{enumerate}
    \item \emph{A Realistic RAG Framework for Indian Courts:} We present \texttt{NyayaRAG}, a novel framework that emulates Indian common law decision-making by incorporating not only facts but also retrieved legal statutes and precedents.
    
    \item \emph{Retrieval-Augmented Pipelines with Structured Inputs:} We construct modular pipelines representing different combinations of factual, statutory, and precedent-based inputs to understand their individual and combined contributions to model performance.


    \item \emph{Simulating Common Law Reasoning with LLMs:} We show that LLMs guided by RAG and factual grounding can produce legally faithful explanations that are aligned with how real-world decisions are made under common law systems.
\end{enumerate}

Our work moves beyond fact-only or self-contained models by replicating a more faithful legal reasoning pipeline aligned with Indian jurisprudence. We hope that \texttt{NyayaRAG} opens new directions for building interpretable, retrieval-aware AI systems in legal settings, particularly in resource-constrained yet precedent-driven judicial systems like India's. We have shared our dataset, code, and RAG-based pipeline through a GitHub repository\footnote{\href{https://github.com/ShubhamKumarNigam/NyayaRAG}{https://github.com/ShubhamKumarNigam/NyayaRAG}}.

%% file: related_work.tex
\section{Related Work}
\label{sec:related-work}
\begin{figure*}[t] 
    \centering 
    \vspace*{-2mm}
    \includegraphics[width=0.75\linewidth]{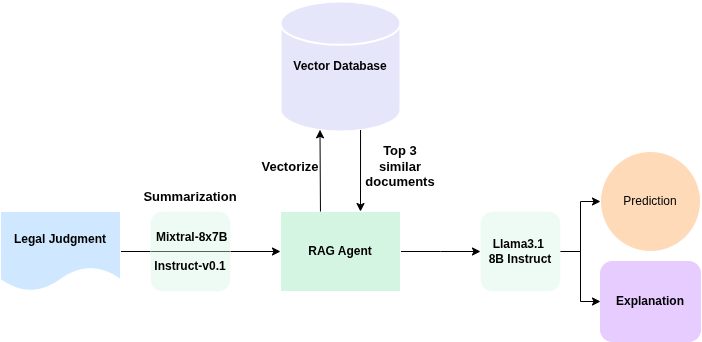} 
    \vspace*{-2mm}
    \caption{Illustration of our Legal Judgment Prediction framework using RAG. The input legal judgment is first summarized; a RAG agent retrieves top-3 relevant documents from a vector database; and an instruction-tuned LLM (e.g., LLaMA-3.1 8B Instruct) generates the final prediction and explanation.}
    \label{fig:task-framework}
\end{figure*}
Recent advances in natural language processing (NLP) and large language models (LLMs) have transformed knowledge-intensive tasks, including question answering and decision support. Transformer-based architectures such as BERT \citep{devlin2018bert}, GPT \citep{radford2019language}, and their instruction-tuned successors demonstrate strong multi-hop reasoning and contextual generation abilities. Retrieval-Augmented Generation (RAG) further improves factual grounding and reduces hallucinations by combining external document retrieval with generative models \citep{han2024rag,hei2024dr}.

The domain of Legal Judgment Prediction (LJP) has evolved from outcome classification toward explanation generation and realistic reasoning. Foundational work explored case outcome prediction in ECHR and Chinese datasets \citep{aletras2016predicting, chalkidis2019neural, xiao2018cail2018}, inspiring subsequent benchmarks such as CAIL2018 and ECHR-CASES. Recent approaches have incorporated statutes, reasoning structures, and event-based features \citep{feng-etal-2022-legal, feng2023criminal}.

Within the Indian legal ecosystem, the ILDC corpus \citep{malik-etal-2021-ildc} laid the foundation for Court Judgment Prediction and Explanation (CJPE). This has been expanded in PredEx \citep{nigam2024legaljudgmentreimaginedpredex}, NyayaAnumana \citep{nigam2024nyayaanumana}, LegalSeg \citep{nigam2025legalseg}, TathyaNyaya \citep{nigam2025tathyanyaya}, and IBPS for bail prediction \citep{srivastava2025ibps}, which collectively focus on segmentation, factual reasoning, and judgment outcome explanation. Efforts such as \citep{nigam2024rethinking,10.1145/3632754.3632765} emphasize fact-based LJP as a realistic alternative to full-text reasoning, aligning predictions more closely with early-phase decision-making. Other works explore semantic segmentation \citep{malik2021semantic}, legal QA \citep{nigam2023legal}, and structured document generation \citep{nigam2025structured}. These resources and models highlight the significance of domain-specific adaptation for Indian law, an underrepresented common law system.

Precedent integration is central to common law reasoning. Early precedent-aware models such as \citep{zhao-etal-2018-learning} focused on Chinese law, while recent works like PLJP \citep{wu2023precedent}, LexKeyPlan \citep{santosh2025lexkeyplan}, and precedent-enhanced ECHR prediction \citep{santosh2024incorporating} examine retrieval of historical judgments to guide predictions. These align with broader multilingual and cross-jurisdictional LJP studies \citep{niklaus2021swiss, kapoor-etal-2022-hldc}. While these efforts highlight the role of past cases, their settings differ significantly from India’s unique common law context where both statutory law and judicial precedents guide reasoning.

Retrieval-augmented pipelines are increasingly applied to legal NLP. LegalBench-RAG \citep{pipitone2024legalbench} and CLERC \citep{hou2024clerc} introduced benchmarks for retrieval-augmented legal reasoning. Specialized systems like CBR-RAG \citep{wiratunga2024cbr}, Graph-RAG for norms \citep{de2025graph}, and LexKeyPlan \citep{santosh2025lexkeyplan} explore structured retrieval, case-based reasoning, and keyphrase-driven planning. Applications range from legal assistants and dispute resolution \citep{rafat2024ai,10887211} to federated secure access architectures \citep{amato2024optimizing}. These works demonstrate the value of augmenting generative reasoning with statutory and precedent-based retrieval but primarily address civil law or international corpora.

Our work contributes to this line by contextualizing RAG for the Indian common law system, simulating how judges reason with case facts, statutory provisions, and precedents. While prior RAG frameworks in law \citep{wiratunga2024cbr, hou2024clerc, wu2023precedent, santosh2024incorporating} establish the feasibility of retrieval-augmented pipelines, they have not been applied to Indian courts. By integrating factual, statutory, and precedential inputs in a unified framework, NyayaRAG extends the landscape of LJP toward realistic, explainable, and domain-grounded legal tasks.

%% file: task_description.tex
\section{Task Description}
\label{sec:task_description}

India’s judicial system operates within the common law framework, where judges deliberate cases based on three fundamental pillars: (i)~the factual context of the case, (ii)~applicable statutory provisions, and (iii)~relevant judicial precedents. Our task is designed to simulate such realistic legal decision-making by leveraging Retrieval-Augmented Generation (RAG), enabling models to access external legal knowledge during inference.

Figure~\ref{fig:task-framework} illustrates our Legal Judgment Prediction (LJP) pipeline enhanced with RAG. The process begins with a full legal judgment document, which undergoes summarization to reduce its length and retain essential factual meaning. This is necessary because legal judgments tend to be long, and appending retrieved knowledge further increases the input size. Given limited model capacity and computational resources, we employ a summarization step (using \texttt{Mixtral-8x7B-Instruct-v0.1}) to create a condensed representation of both the input case and the retrieved legal context.

\paragraph{Prediction Task:} Based on the summarized factual description and the retrieved top-$k$ similar legal documents (statutes or precedents), the model predicts the likely court judgment. The prediction label \(y \in \{0, 1\}\) indicates whether the appeal is fully rejected ($0$) or fully/partially accepted ($1$). This binary framing captures the most common forms of judicial decisions in Indian appellate courts.

\paragraph{Explanation Task:} Alongside the decision, the model is also required to generate an explanation that justifies its output. This explanation should logically incorporate the facts, cited statutes, and relevant precedents retrieved during the RAG process. This step emulates how judges provide reasoned opinions in written judgments.

By structuring the LJP task in this manner, and by summarizing long documents and integrating retrieval-based augmentation, we study the effectiveness of RAG in producing judgments that are both faithful to legal reasoning and grounded in precedent and statute. The overall framework allows us to approximate real-world decision-making environments within Indian courtrooms.

%% file: dataset.tex
\section{Dataset}
\label{sec:dataset}
Our dataset is designed to simulate realistic court decision-making in the Indian legal context, incorporating facts, statutes, and precedent, which are essential elements under the common law framework. This dataset enables exploration of Legal Judgment Prediction (LJP) in a Retrieval-Augmented Generation (RAG) setup.

\subsection{Dataset Compilation}

We curated a large-scale dataset consisting of 56,387 Supreme Court of India (SCI) case documents up to April 2024, sourced from IndianKanoon\footnote{\url{https://indiankanoon.org/}}, a trusted legal search engine. The website provides structural tags for various judgment components (e.g., facts, issues, arguments), which allowed for clean and structured scraping. These documents serve as the foundation for our summarization, retrieval, and reasoning experiments.

\subsection{Single versus Multi Partitioning}

Another challenge in the SCI proceedings is the presence of cases containing multiple petitions under a single case identifier, where each petition may lead to distinct outcomes. To handle this, we partition the dataset into two subsets. The \textit{single} partition represents binary classification tasks (\texttt{accept}/\texttt{reject}) and includes cases with a single petition or multiple petitions that share the same outcome. The \textit{multi} partition
contains cases with multiple petitions leading to different decisions, introducing a more complex multi-label prediction setting (e.g., partially accepted outcomes). For these cases, if even one appeal within the group is accepted, the case is assigned the label \texttt{accepted}. Predicting mixed outcomes remains challenging and will be further explored in future work.

\subsection{Dataset Composition}

The corpus supports multiple downstream pipelines, each focusing on specific judgment elements or legal context. Table~\ref{tab:test-cases} presents key statistics across different configurations.
An example case is shown in Table~\ref{case-example} in Appendix.

\subsubsection{Case Text}

Each judgment includes complete narrative content such as factual background, party arguments, legal issues, reasoning, and verdict. Due to length constraints exceeding model context windows, we summarized these documents using \texttt{Mixtral-8x7B-Instruct-v0.1} \citep{Jiang2024MixtralOE}, which supports up to 32k tokens. The summarization preserved critical legal elements through carefully designed prompts (see Table~\ref{tab:Instruction-sets-summarize}).

\begin{table}[t]
\centering
\resizebox{\columnwidth}{!}{
\begin{tabular}{lrrr}
\hline
\textbf{Dataset}  & \textbf{\#Documents} & \textbf{Avg. Length} & \textbf{Max} \\ \hline
\textbf{SCI (Full)} & 56,387 & 3,495 & 401,985  \\
\textbf{Summarized Single} & 4,962 & 302 & 875  \\
\textbf{Summarized Multi}  & 4,930 & 300 & 879  \\
\textbf{Sections} & 29,858 & 257 & 27,553  \\ \hline
\end{tabular}
}
\vspace*{-2mm}
\caption{NyayaRAG data statistics}
\label{tab:test-cases}
\end{table}

\begin{table*}[t]
\centering
\begin{tabular}{p{0.90\textwidth}}
\toprule
\textbf{Summarization Prompt} \\
\midrule
\footnotesize
The text is regarding a court judgment for a specific case. Summarize it into 1000 tokens but more than 700 tokens. The summarization should highlight the Facts, Issues, Statutes, Ratio of the decision, Ruling by Present Court (Decision), and a Conclusion. \\
\bottomrule
\end{tabular}
\vspace*{-2mm}
\caption{Instruction prompt used with \texttt{Mixtral-8x7B-Instruct-v0.1} for summarizing legal judgments}
\label{tab:Instruction-sets-summarize}
\end{table*}

\subsubsection{Precedents}

From each judgment, cited precedents were extracted using metadata tags provided by IndianKanoon. These citations represent explicit legal reasoning and are retained for use during inference to replicate how courts consider prior judgments.

\subsubsection{Statutes}

Statutory references were also programmatically extracted, including citations to laws like the Indian Penal Code and the Constitution of India. Where statute sections exceeded length limits, they were summarized using the same LLM pipeline. Only statutes directly cited in the respective cases were retained, ensuring relevance.

\subsubsection{Previous Similar Cases}
To simulate implicit precedent-based reasoning, we employed semantic similarity retrieval to identify relevant previous cases beyond explicit citations:

\begin{itemize}
    \item \textbf{Corpus Vectorization:} All 56,387 documents were embedded into dense vector representations using the \texttt{all-MiniLM-L6-v2} sentence transformer.
    \item \textbf{Target Encoding:} The 5,000 selected training samples were vectorized similarly.
    \item \textbf{Top-k Retrieval:} Using \texttt{ChromaDB}, we retrieved the top-3 most semantically similar cases for each document based on cosine similarity.
    \item \textbf{Augmentation:} Retrieved cases were appended to the factual input to form the ``casetext + previous similar cases'' input during model inference.
\end{itemize}

This retrieval step enriches context with precedents that are semantically close; even if not cited, they enhance the legal realism of our setup.

\subsubsection{Facts}
We separately extracted the factual portions of all 56,387 judgments. These include background information, chronological events, and party narratives, excluding legal reasoning. These fact-only subsets were used to simulate realistic courtroom scenarios where judges primarily rely on facts, relevant law, and precedent for decision-making.

Overall, our dataset is uniquely structured to test legal decision-making under realistic constraints, aligning with the Indian legal system's reliance on factual narratives, statutory frameworks, precedents, and prior rulings.

%% file: methodology.tex
\section{Methodology}
\label{sec:methodology}
To simulate realistic judgment prediction and evaluate the role of RAG in enhancing legal decision-making, we design a modular experimental setup. This setup explores how different types of legal information, such as factual summaries, statutes, and precedents affect model performance on the dual tasks of prediction and explanation. To ensure reproducibility and transparency, we detail the full experimental setup, including model configurations, training routines, and task-specific hyperparameters. The details are mentioned in Appendix~\ref{sec:Experimental-setup}. This includes separate sections for explanation generation (summarization) and legal judgment prediction tasks, outlining all relevant decoding strategies, optimization settings, and dataset splits used across our pipeline variants.

\subsection{Pipeline Construction}
To systematically evaluate the impact of legal knowledge sources, we constructed multiple input pipelines using combinations of the dataset components described in Section~\ref{sec:dataset}. Each pipeline configuration represents a distinct input scenario reflecting different degrees of legal context and retrieval augmentation. These pipelines are as follows:

\begin{itemize}
    \item \textbf{CaseText Only:} Includes only the summarized version of the full case judgment, which contains factual background, arguments, and reasoning.

    \item \textbf{CaseText + Statutes:} Appends summarized statutory references cited in the judgment to the case text, simulating scenarios where relevant laws are explicitly considered.

    \item \textbf{CaseText + Precedents:} Incorporates prior cited judgments mentioned in the original case, representing explicitly relied-upon precedents.

    \item \textbf{CaseText + Previous Similar Cases:} Adds top-3 semantically similar past judgments (retrieved via ChromaDB using \texttt{all-MiniLM-L6-v2} embeddings), allowing the model to learn from precedents not explicitly cited.

    \item \textbf{CaseText + Statutes + Precedents:} A comprehensive legal input pipeline combining the full judgment summary, statutes, and cited prior judgments.

    \item \textbf{Facts Only:} A minimal pipeline containing only the factual summary, excluding all legal reasoning and verdicts. This setup evaluates whether a model can infer judgments from facts alone.

    \item \textbf{Facts + Statutes + Precedents:} Combines factual input with statutory and precedent context to simulate realistic courtroom conditions where judges rely on facts, applicable law, and relevant past cases.
\end{itemize}

This modular design enables granular control over input features and facilitates direct comparison of how each knowledge source contributes to judgment prediction and explanation generation.

\subsection{Prompt Design}
\label{subsec:prompts}
To ensure consistency and interpretability across all pipelines, we used fixed instruction prompts with minor variations depending on the available contextual inputs (e.g., facts only vs. facts + law + precedent). These prompts guide the model in producing both binary predictions and natural language explanations. Prompts were structured to reflect real judicial inquiry formats, aligning with the instruction-following capabilities of modern LLMs. Full prompt templates are listed in Appendix Table~\ref{tab:judgment_prediction_prompts_few}, along with prediction examples.

\subsection{Inference Setup}
\label{subsec:inference}
We use the \texttt{LLaMA-3.1 8B Instruct}~\cite{dubey2024llama} model for all experiments in a few-shot prompt setup. Each input sequence, composed according to one of the pipeline templates, is paired with a relevant prompt that asks to output:

\begin{itemize}
    \item A binary judgment prediction: {0} (appeal rejected) or {1} (appeal fully/partially accepted)
    \item A justification: a coherent explanation based on legal facts, statutes, and precedent
\end{itemize}

The model is explicitly instructed to reason with the provided information and emulate judicial writing. Retrieved knowledge (via RAG) is included in-context to enhance legal reasoning while minimizing hallucinations.

This experimental design allows us to evaluate the effectiveness of legal retrieval and summarization under realistic judicial decision-making constraints in the Indian common law setting.

%% file: evaluation_metrics.tex
\section{Evaluation Metrics}
\label{sec:performance_metrics}
\begin{table*}[t]
\centering
\resizebox{0.70\textwidth}{!}{
\begin{tabular}{l lcccc}
\toprule
\textbf{Pipeline} & \textbf{Partition} & \textbf{Accuracy} & \textbf{Precision} & \textbf{Recall} & \textbf{F1-score} \\
\midrule
\multirow{2}{*}
{CaseText Only}
  & Single & 62.27 & 33.50 & 30.88 & 29.45 \\
  & Multi  & 53.10 & 25.26 & 23.95 & 20.81 \\
\midrule
\multirow{2}{*}
{CaseText + Statutes}
  & Single & \textbf{67.07} & 45.29 & 44.55 & 44.32 \\
  & Multi  & 60.36 & \textbf{64.22} & \textbf{64.04} & 60.35 \\
\midrule
\multirow{2}{*}
{CaseText + Precedents}
  & Single & 61.73 & 41.92 & 41.35 & 40.81 \\
  & Multi  & 57.53 & 61.34 & 61.19 & 57.53 \\
\midrule
\multirow{2}{*}
{CaseText + Previous Similar Cases}
  & Single & 57.53 & \textbf{61.34} & \textbf{61.19} & \textbf{57.53} \\
  & Multi  & 61.73 & 41.92 & 41.35 & 57.53 \\
\midrule
\multirow{2}{*}
{CaseText + Statutes + Precedents}
  & Single & 64.71 & 43.50 & 42.98 & 42.78 \\
  & Multi  & \textbf{65.86} & 63.94 & 63.99 & \textbf{63.96} \\
\midrule
\multirow{2}{*}
{CaseFacts Only}
  & Single & 51.13 & 51.36 & 51.30 & 50.68 \\
  & Multi  & 53.71 & 51.18 & 51.18 & 51.18 \\
\midrule
\multirow{2}{*}
{Facts + Statutes + Precedents}
  & Single & 50.58 & 33.57 & 33.56 & 33.24 \\
  & Multi  & 52.57 & 52.01 & 52.01 & 52.01 \\
\bottomrule
\end{tabular}
}
\vspace*{-2mm}
\caption{Performance of various pipelines on Binary and Multi-label LJP task (best results in bold)}
\label{tab:prediction_pipeline_results}
\end{table*}

To evaluate the effectiveness of our Retrieval-Augmented Legal Judgment Prediction framework, we adopt a comprehensive set of metrics covering both classification accuracy and explanation quality. The evaluation is conducted on two fronts: the judgment prediction task and the explanation generation task. These metrics are selected to ensure a holistic assessment of model performance in the legal domain. We report Macro Precision, Macro Recall, Macro F1, and Accuracy for judgment prediction, and we use both quantitative and qualitative methods to evaluate the quality of explanations generated by the model.

\begin{enumerate}
    \item \textbf{Lexical-based Evaluation:} We utilized standard lexical similarity metrics, including Rouge-L~\cite{lin-2004-rouge}, BLEU \cite{papineni-etal-2002-bleu}, and METEOR \cite{banerjee-lavie-2005-meteor}. These metrics measure the overlap and order of words between the generated explanations and the reference texts, providing a quantitative assessment of the lexical accuracy of the model outputs.

    \item \textbf{Semantic Similarity-based Evaluation:} To capture the semantic quality of the generated explanations, we employed BERTScore \cite{BERTScore}, which measures the semantic similarity between the generated text and the reference explanations. Additionally, we used BLANC \cite{blanc}, a metric that estimates the quality of generated text without a gold standard, to evaluate the model's ability to produce semantically meaningful and contextually relevant explanations.

    \item \textbf{LLM-based Evaluation (LLM-as-a-Judge):} To complement traditional metrics, we incorporate an automatic evaluation strategy that uses large language models themselves as evaluators, commonly referred to as \textit{LLM-as-a-Judge}. This evaluation is crucial for assessing structured argumentation and legal correctness in a format aligned with expert judicial reasoning. We adopt G-Eval~\cite{liu-etal-2023-g}, a GPT-4-based evaluation framework tailored for natural language generation tasks. G-Eval leverages chain-of-thought prompting and structured scoring to assess explanations along three key criteria: \textit{factual accuracy}, \textit{completeness \& coverage}, and \textit{clarity \& coherence}. Each generated legal explanation is scored on a scale from 1 to 10 based on how well it aligns with the expected content and a reference document. The exact prompt format used for evaluation is shown in Appendix Table~\ref{tab:G-Eval Prompt}. For our experiments, we use the GPT-4o-mini model to generate reliable scores without manual intervention. This setup provides an interpretable, unified judgment metric that captures legal soundness, completeness of reasoning, and logical coherence, beyond what traditional similarity-based metrics can offer.


    \item \textbf{Expert Evaluation:} To validate the interpretability and legal soundness of the model-generated explanations, we conduct an expert evaluation involving legal professionals. They rate a representative subset of the generated outputs on a 1–10 Likert scale across three criteria: factual accuracy, legal relevance, and completeness of reasoning. A score of 1 denotes a poor or misleading explanation, while a 10 reflects high legal fidelity and argumentative soundness. This evaluation provides critical insights beyond automated metrics.

    \item \textbf{Inter-Annotator Agreement (IAA):} To further ensure the reliability and consistency of expert judgments, we computed multiple standard inter-rater agreement metrics. Specifically, Fleiss’ Kappa~\cite{fleiss1971measuring} was used to evaluate the overall agreement among multiple raters, Cohen’s Kappa~\cite{cohen1960coefficient} captured pairwise agreement while adjusting for chance, and the Intraclass Correlation Coefficient (ICC)~\cite{shrout1979intraclass} measured the reliability of continuous ratings across raters. In addition, Krippendorff’s Alpha~\cite{krippendorff2018content} provided a robust measure suitable for ordinal scales and missing data, while the Pearson Correlation Coefficient~\cite{benesty2009pearson} quantified the linear consistency of expert scores. The results demonstrated substantial agreement across different measures, reinforcing the credibility of the evaluation and lending strong support to the robustness of our findings.
\end{enumerate}

Together, these metrics provide a comprehensive, multi-perspective evaluation of our system’s ability to both predict judicial outcomes and generate legally coherent, interpretable rationales grounded in retrieved context.


\begin{table*}[t]
\centering
\resizebox{0.80\linewidth}{!}{
\begin{tabular}{lccccccc}
\toprule
\textbf{Pipelines} & \textbf{RL} & \textbf{BLEU} & \textbf{METEOR} & \textbf{BERTScore} & \textbf{BLANC} & \textbf{G-Eval} & \textbf{Expert Score} \\
\midrule
\multicolumn{8}{c}{\textbf{Single Partition}} \\
\midrule
CaseText Only & 0.16 & 0.03 & 0.18 & 0.52 & 0.08 & 4.17 & 5.2 \\
CaseText + Statutes & \textbf{0.17} & \textbf{0.03} & \textbf{0.20} & \textbf{0.53} & \textbf{0.09} & \textbf{4.21} & \textbf{5.5} \\
CaseText + Precedents & 0.16 & 0.03 & 0.19 & 0.51 & 0.08 & 3.45 & 4.6 \\
CaseText + Previous Similar Cases & 0.16 & 0.03 & 0.20 & 0.52 & 0.08 & 3.72 & 4.9 \\
CaseText + Statutes + Precedents & 0.16 & 0.03 & 0.19 & 0.52 & 0.08 & 4.11 & 5.4 \\
CaseFacts Only & 0.16 & 0.02 & 0.18 & 0.52 & 0.06 & 3.53 & 4.5 \\
Facts + Statutes + Precedents & 0.16 & 0.02 & 0.18 & 0.51 & 0.06 & 2.97 & 3.9 \\
\midrule
\multicolumn{8}{c}{\textbf{Multi Partition}} \\
\midrule
CaseText Only & 0.16 & 0.03 & 0.18 & 0.52 & 0.08 & 4.00 & 5.0 \\
CaseText + Statutes & \textbf{0.17} & \textbf{0.03} & \textbf{0.20} & \textbf{0.53} & \textbf{0.09} & \textbf{4.10} & \textbf{5.3} \\
CaseText + Precedents & 0.16 & 0.03 & 0.20 & 0.53 & 0.09 & 3.41 & 4.4 \\
CaseText + Previous Similar Cases & 0.16 & 0.03 & 0.19 & 0.52 & 0.08 & 3.67 & 4.7 \\
CaseText + Statutes + Precedents & 0.16 & 0.03 & 0.20 & 0.53 & 0.09 & 3.92 & 5.2 \\
CaseFacts Only & 0.15 & 0.02 & 0.17 & 0.52 & 0.08 & 3.74 & 4.6 \\
Facts + Statutes + Precedents & 0.15 & 0.02 & 0.19 & 0.52 & 0.07 & 3.08 & 4.1 \\
\bottomrule
\end{tabular}
}
\vspace*{-2mm}
\caption{Comparison of explanation generation across different legal context pipelines}
\label{tab:explanation_pipeline_results}
\end{table*}

%% file: results_and_analysis.tex
\section{Results and Analysis}
\label{sec:results_analysis}

We conducted extensive evaluations across multiple pipeline configurations to study the impact of different legal information components on both judgment prediction and explanation quality. Tables~\ref{tab:prediction_pipeline_results} and~\ref{tab:explanation_pipeline_results} summarize the model's performance across these configurations for binary and multi-label settings.

\subsection{Judgment Prediction Performance}

As shown in Table~\ref{tab:prediction_pipeline_results}, the pipeline combining \textit{CaseText + Statutes} achieved the highest accuracy in the single-label setting. This suggests that legal statutes provide substantial contextual cues for the model to infer the likely decision. In contrast, \textit{CaseText Only} achieved 62.27\%, highlighting the importance of augmenting case narratives with applicable laws. Interestingly, the \textit{CaseText + Previous Similar Cases} pipeline showed the highest precision, recall, and F1-score in the single-label case, indicating that semantically retrieved precedents, despite not being explicitly cited, help the model align with actual judicial outcomes.

In the multi-label setting, the best accuracy was observed for the \textit{CaseText + Statutes + Precedents} pipeline. This comprehensive context provides the model with structured legal knowledge, improving generalization across different outcome labels. Conversely, the \textit{Facts Only} pipeline performed worst overall, reaffirming that factual narratives alone, without legal context, are insufficient for reliably predicting legal outcomes. The poor performance of the \textit{Facts + Statutes + Precedents} pipeline in the single-label setting suggests that factual sections lack the interpretive cues that full case texts offer when combined with legal references.

\subsection{Explanation Generation Quality}

Table~\ref{tab:explanation_pipeline_results}presents the results of explanation evaluation using a diverse set of metrics, including both automatic lexical and semantic metrics (ROUGE, BLEU, METEOR, BERTScore, BLANC) and a large language model-based evaluation (G-Eval). Across both single and multi-label setups, the \textit{CaseText + Statutes} pipeline consistently outperformed all other configurations. In the single-label setting, it achieved the highest scores across key dimensions, substantially outperforming the \textit{CaseText Only} baseline. This result underscores the critical role of statutory references in enhancing both the factual alignment and interpretability of model-generated legal explanations.

Interestingly, while the \textit{CaseText + Previous Similar Cases} pipeline yielded strong lexical overlap (e.g., top ROUGE-L in the unabridged version), it lagged behind the statute-enhanced pipeline in metrics that assess semantic and contextual alignment, such as G-Eval and BLANC. This indicates that while similar cases might help the model replicate surface-level language, they may not consistently offer legally grounded or complete reasoning. Meanwhile, the \textit{CaseText + Statutes + Precedents} pipeline also performed competitively, suggesting that combining structured legal references with precedent data can lead to balanced and high-quality explanations.

In contrast, configurations that relied solely on factual narratives (\textit{CaseFacts Only} and \textit{Facts + Statutes + Precedents}) exhibited comparatively poor performance across all evaluation metrics. For example, the \textit{Facts + Statutes + Precedents} pipeline recorded a G-Eval score as low in the single-label setting. This reinforces the notion that factual descriptions, while essential, are insufficient for constructing legally persuasive rationales. The absence of structured legal arguments, statutory alignment, or precedent citation in these setups appears to undermine their explanatory effectiveness.

\begin{table*}[t]
\centering
\resizebox{0.70\textwidth}{!}{
\begin{tabular}{lccccc}
\toprule
\textbf{Pipelines} & \textbf{Fleiss' $\kappa$} & \textbf{Cohen's $\kappa$} & \textbf{ICC} & \textbf{Kripp. $\alpha$} & \textbf{Pearson Corr.} \\
\midrule
\multicolumn{6}{c}{\textbf{Single Partition}} \\
\midrule
CaseText Only & 0.42 & 0.47 & 0.55 & 0.49 & 0.58 \\
CaseText + Statutes & 0.51 & 0.55 & 0.61 & 0.57 & 0.65 \\
CaseText + Precedents & 0.37 & 0.42 & 0.50 & 0.45 & 0.52 \\
CaseText + Previous Similar Cases & 0.41 & 0.45 & 0.54 & 0.47 & 0.56 \\
CaseText + Statutes + Precedents & 0.49 & 0.52 & 0.59 & 0.54 & 0.62 \\
CaseFacts Only & 0.34 & 0.39 & 0.47 & 0.43 & 0.49 \\
Facts + Statutes + Precedents & 0.29 & 0.34 & 0.42 & 0.38 & 0.44 \\
\midrule
\multicolumn{6}{c}{\textbf{Multi Partition}} \\
\midrule
CaseText Only & 0.40 & 0.45 & 0.53 & 0.48 & 0.57 \\
CaseText + Statutes & 0.50 & 0.54 & 0.60 & 0.56 & 0.64 \\
CaseText + Precedents & 0.36 & 0.41 & 0.49 & 0.44 & 0.51 \\
CaseText + Previous Similar Cases & 0.39 & 0.44 & 0.52 & 0.46 & 0.55 \\
CaseText + Statutes + Precedents & 0.47 & 0.51 & 0.58 & 0.52 & 0.60 \\
CaseFacts Only & 0.32 & 0.37 & 0.46 & 0.41 & 0.48 \\
Facts + Statutes + Precedents & 0.28 & 0.33 & 0.41 & 0.36 & 0.43 \\
\bottomrule
\end{tabular}
}
\vspace*{-2mm}
\caption{Inter-Annotator Agreement (IAA) statistics for expert evaluation of generated legal explanations across different pipeline settings for both Single and Multi partitions}
\label{tab:iaa_expert_scores}
\end{table*}

\paragraph{Expert Evaluation:}
The results of this evaluation aligned closely with the trends observed in automatic metrics (Table~\ref{tab:explanation_pipeline_results}). Pipelines enriched with statutory provisions, particularly \textit{CaseText + Statutes}, consistently received the highest expert ratings, highlighting the value of legal grounding in improving explanation quality. In contrast, pipelines relying solely on factual input, such as \textit{CaseFacts Only}, obtained the lowest expert scores, underscoring their lack of interpretive depth.

To assess the reliability of these expert judgments, we performed a detailed Inter-Annotator Agreement (IAA) analysis across multiple evaluation dimensions, Table~\ref{tab:iaa_expert_scores}. The IAA results revealed moderate to substantial agreement among annotators. For the \textit{Single} partition, pipelines such as \textit{CaseText + Statutes} and \textit{CaseText + Statutes + Precedents} achieved higher consistency (Fleiss’ $\kappa > 0.49$, ICC $\approx 0.60$), reflecting strong consensus on their explanation quality. In contrast, fact-only or noisy-input pipelines showed weaker agreement (Fleiss’ $\kappa < 0.35$, ICC $< 0.50$), suggesting greater variability in expert perception of their outputs.

The {Multi} partition exhibited slightly lower agreement overall, likely due to the additional complexity of multi-label judgments. Nevertheless, richer legal context pipelines again demonstrated higher consistency among experts compared to fact-only inputs. These findings not only validate the robustness of our expert evaluation protocol but also corroborate trends from lexical, semantic, and LLM-based evaluations.

Taken together, the results emphasize that Retrieval-Augmented Generation, when paired with structured legal inputs such as statutes and precedents, produces explanations that are more accurate, interpretable, and legally coherent. The combination of automatic metrics, LLM-based judgment, and human expert ratings provides a multifaceted and credible framework for assessing explanation quality in legal judgment prediction.

%% file: Ablation_study.tex
\section{Ablation Study: Understanding the Role of Legal Context Components}
\label{sec:ablation_study}

To assess the individual contribution of each legal context component, factual narratives, statutory provisions, cited precedents, and semantically similar past cases, we perform an ablation study by systematically removing or altering these inputs across pipeline configurations. This study highlights how each component affects prediction accuracy and explanation quality, as reported in Tables~\ref{tab:prediction_pipeline_results} and~\ref{tab:explanation_pipeline_results}.

\paragraph{Impact on Judgment Prediction:}
The \texttt{CaseText + Statutes + Precedents} pipeline serves as the most comprehensive baseline. Removing statutory references (i.e., \texttt{CaseText + Precedents}) leads to a noticeable drop in F1-score (from 63.96 to 57.53 in the multi-label setting), indicating that legal provisions provide structured grounding essential for accurate predictions. Similarly, eliminating precedents (i.e., \texttt{CaseText + Statutes}) also reduces performance, though the drop is less steep, suggesting complementary roles of statutes and precedents. Pipelines relying solely on factual case narratives (e.g., \texttt{CaseFacts Only}) perform the worst, reaffirming that factual information alone is insufficient for robust legal outcome prediction.

\paragraph{Impact on Explanation Quality:}
A similar pattern emerges in explanation generation. The \texttt{CaseText + Statutes} pipeline consistently outperforms others across ROUGE, BLEU, METEOR, BERTScore, and G-Eval metrics, underscoring the importance of grounding explanations in explicit statutory language. When only precedents are added (without statutes), as in \texttt{CaseText + Precedents}, explanation scores drop significantly (e.g., G-Eval: 4.21 to 3.45 in the single-label case). The worst-performing setup is \texttt{Facts + Statutes + Precedents}, highlighting that factual inputs, even when supplemented with legal references, do not suffice for generating coherent and persuasive explanations if the core case context is missing.

\paragraph{Insights:}
These findings validate the design choices in \texttt{NyayaRAG}, where integrating factual case text with statutory and precedential knowledge mimics real-world judicial reasoning. Statutory references provide normative structure, while precedents offer context-specific analogies. Their absence not only reduces predictive performance but also degrades the factuality, clarity, and legal coherence of the generated explanations.

This ablation analysis also offers practical guidance: for retrieval-augmented systems deployed in legal contexts, careful curation and combination of retrieved statutes and relevant precedents are critical to ensure trustworthy outputs.

%% file: conculsion.tex
\section{Conclusions and Future Scope}
\label{sec:conclusion_future_scope}

This paper introduced \texttt{NyayaRAG}, a Retrieval-Augmented Generation framework tailored for realistic legal judgment prediction and explanation in the Indian common law system. By combining factual case details with retrieved statutory provisions and relevant precedents, our approach mirrors judicial reasoning more closely than prior methods that rely solely on the case text.
Empirical results across prediction and explanation tasks confirm that structured legal retrieval enhances both outcome accuracy and interpretability. Pipelines enriched with statutes and precedents consistently outperformed baselines, as validated by lexical, semantic, and LLM-based (G-Eval) metrics, as well as expert feedback.

Future directions include extending to hierarchical verdict structures, integrating symbolic or graph-based retrieval, modeling temporal precedent evolution, and leveraging human-in-the-loop mechanisms. \texttt{NyayaRAG} marks a step toward court-aligned, explainable legal AI and sets the foundation for future research in retrieval-enhanced legal systems within under-represented jurisdictions.

%% file: acknowledgement.tex
\section*{Acknowledgements}

We would like to express our sincere gratitude to the anonymous reviewers for their constructive feedback and valuable suggestions, which greatly improved the quality of this paper. We also thank the student research assistants and legal experts from various law colleges for their dedicated efforts in annotation and expert evaluation. Their contributions were instrumental in ensuring the quality and reliability of the dataset and evaluation process.
We gratefully acknowledge the support of BharatGen, India, for providing access to computational resources and hardware infrastructure used in this research. Their assistance played a vital role in model training and large-scale experimentation. The majority of this work was conducted while the first author was affiliated with the Indian Institute of Technology Kanpur. The author is currently affiliated with the University of Birmingham Dubai.

%% file: limitation.tex
\section*{Limitations}
\label{sec:limitations}

While \texttt{NyayaRAG} marks a significant advance in realistic legal judgment prediction under the Indian common law framework, several limitations merit further attention.

First, although Retrieval-Augmented Generation (RAG) helps reduce hallucinations by grounding outputs in retrieved legal documents, it does not fully eliminate factual or interpretive inaccuracies. In sensitive domains such as law, even rare errors in reasoning or justification may raise concerns about reliability and accountability.

Second, the current framework supports binary and multi-label outcome structures but does not yet handle the full spectrum of legal verdicts, such as hierarchical or multi-class decisions involving complex legal provisions. Expanding to richer verdict taxonomies would enable broader applicability and deeper case understanding.

Third, \texttt{NyayaRAG} assumes the availability of clean, well-structured legal documents and relies on summarization pipelines to manage input length. However, real-world legal texts often contain noise, OCR errors, or inconsistent formatting. Although summarization aids conciseness, it may inadvertently omit subtle legal nuances that affect judgment outcomes or explanation quality.

Finally, due to computational resource constraints, the current system utilizes instruction-tuned LLMs guided by domain-specific prompts rather than fully fine-tuning on large-scale Indian legal corpora. While prompt-based tuning remains efficient and modular, fine-tuning on in-domain legal texts could further enhance model fidelity and domain alignment.

Despite these limitations, \texttt{NyayaRAG} provides a robust and interpretable foundation for judgment prediction and explanation, supported by both automatic and expert evaluations. Future work that addresses these constraints, particularly hierarchical decision modeling and domain-specific fine-tuning, will further strengthen the framework’s legal relevance and practical deployment potential.

%% file: ethics.tex
\section*{Ethics Statement}

This research adheres to established ethical standards for conducting work in high-stakes domains such as law. The legal documents used in our study were sourced from IndianKanoon (\url{https://indiankanoon.org/}), a publicly available repository of Indian court judgments. All documents are in the public domain and do not include sealed cases or personally identifiable sensitive information, ensuring that our use of the data complies with privacy and confidentiality norms.

We emphasize that the proposed \texttt{NyayaRAG} system is developed strictly for academic research purposes to simulate realistic legal reasoning processes. It is not intended for direct deployment in real-world legal settings. The model outputs must not be construed as legal advice, official court predictions, or determinants of legal outcomes. Any downstream use should be performed with oversight by qualified legal professionals. We strongly discourage the use of this system in live legal cases, policymaking, or decisions that may affect individuals' rights without appropriate human-in-the-loop supervision.

As part of our evaluation protocol, we involved domain experts (legal professionals and researchers) to assess the quality and legal coherence of the generated explanations. The evaluation was conducted on a curated subset of samples, and all participating experts were informed of the research objectives and voluntarily participated without any coercion or conflict of interest. No personal data was collected during this process, and all expert feedback was anonymized for analysis.

While we strive to enhance legal interpretability and transparency, we acknowledge that legal documents themselves may reflect systemic biases. Our framework, while replicating judicial reasoning patterns, may inherit such biases from training data. We do not deliberately introduce or amplify such biases, but we recognize the importance of further work in fairness auditing, particularly across litigant identity, socio-demographic markers, and jurisdictional diversity.

%% file: appendix.tex
\section{Experimental Setup and Hyper-parameters}
\label{sec:Experimental-setup}
\subsection{Summarization Hyper-parameters}
To condense lengthy Indian Supreme Court judgments into structured and model-friendly inputs, we employed \texttt{Mixtral-8x7B-Instruct-v0.1}, a mixture-of-experts, instruction-tuned language model developed by Mistral AI. The summarization was conducted in a zero-shot setting using tailored legal prompts that extracted key elements such as facts, statutes, precedents, reasoning, and the final ruling.

The model was accessed via the HuggingFace Transformers interface and run on an NVIDIA A100 GPU with 80GB VRAM. Inputs were truncated to a maximum of 27,000 tokens to comply with the model’s context window. The output length was constrained to between 700 and 1,000 tokens to ensure consistency and legal completeness. A low decoding temperature of 0.2 was used to encourage determinism and factual alignment. These summaries served as inputs to the Retrieval-Augmented Generation (RAG) pipelines used for downstream judgment prediction and explanation.

\subsection{Judgment Prediction Hyper-parameters}
For the legal judgment prediction task, we used the \texttt{LLaMA 3–8B Instruct} model, which supports high-quality reasoning in instruction-following settings. The model was applied in a few-shot prompting setup without any task-specific fine-tuning. Input prompts consisted of structured summaries (produced by Mixtral) along with retrieved statutes and prior similar cases. These inputs followed a consistent legal instruction format to guide the model's prediction and explanation generation.

Inference was performed using the PyTorch backend with HuggingFace Transformers on an NVIDIA A100 GPU (80GB). The model was loaded using \texttt{device\_map=``auto''} for automatic device allocation. We used deterministic generation parameters (temperature = 0.2, top-p = 0.9) and controlled output format to ensure faithful and interpretable outputs. Each output consisted of a binary prediction (\texttt{0} for appeal rejected, \texttt{1} for appeal accepted/partially accepted) followed by a free-text legal explanation. No supervised fine-tuning was used, which allows our framework to be easily adapted to different legal datasets without retraining.

\begin{table*}[ht]
\centering
\tiny
\resizebox{\textwidth}{!}{%
\begin{tabular}{|c|l|}
\hline

\textbf{Pipeline} &
  \multicolumn{1}{c|}{\textbf{Decision and Explanation}} \\ \hline
 
Ground Truth &
  { \begin{tabular}[c]{@{}l@{}}Decision: Rejected. \\Explanation:
  \\1. The parties to these appeals belong to Chokkura Thaliyadath tarwad, a hindu \\family governed by Marumakkathayam system of law as modified by the Madras \\Marumakkathayam Act,1932. They belong to the thavazhi of Cheriyammu Amma. \\Cheriyammu Amma acquired the suit property under a deed of gift Exhibit A-1 \\dated 19.6.1905.\\
  ........................................................\\
  Thus, in both the suits, the essential question that fell for decision, especially in the \\context of the only contention raised before us by learned Senior Counsel Mr. A.S. \\Nambiar, appearing on behalf of the appellant, was whether the transaction Exhibit-\\B-9 could be ignored by the sons of Kunhimalu Amma as a void transaction.", '6. \\Exhibit-B-9 is termed a deed of release\end{tabular}} \\ \hline

Case Text only &
  { \begin{tabular}[c]{@{}l@{}}Decision: Rejected.\\ Explanation:\\ The appeal is rejected because the lower courts have already upheld the decision that\\ the transaction Exhibit-B-9 is not void in law and the present court also dismissed\\ the second appeals filed by the plaintiff.\\
  ........................................................\\
  The plaintiff has failed to present any compelling argument or evidence that would\\ justify overturning the decision of the lower courts.  The appeal is therefore rejected.  \\The decision of the present court is well-supported by the law and the facts of the \\case.  The plaintiff has no valid grounds for appeal.\end{tabular}} \\ \hline

Case Text with Statutes &
  { \begin{tabular}[c]{@{}l@{}}Decision: Rejected \\ Explanation: 
  \\The appeal is rejected because the present court upholds the decision of the lower\\ courts, and the transaction Exhibit-B-9 is not void in law as it is a release of rights \\by all the other members of the thavazhi in favor of the only other member and \\not an assignment of undivided shares. \\
  ........................................................\\
  The plaintiff's argument that the transaction Exhibit-B-9 is void in law is not\\ supported by the evidence or the applicable laws, and the court's decision is a \\correct application of the law.  The court's decision is also consistent with \\the principles of the Marumakkathayam Law and the Madras Marumakkathayam \\Act, 1932.  The facts of the case do not support the plaintiff's argument that the\end{tabular}} \\ \hline

Case Text with Precedents &
  {\begin{tabular}[c]{@{}l@{}}Decision: Rejected \\Explanation:
  \\The appeal is rejected because the transaction Exhibit-B-9 is not void in law as it\\ is a release of rights by all the other members of the thavazhi in favor of the only\\ other member and not an assignment of undivided shares, which is in accordance \\with the Madras Marumakkathayam Act, 1932, and the ratio of the decision.\\
  ........................................................\\
  The ratio of the decision is that the transaction Exhibit-B-9 is not void in law as\\ it is a release of rights by all the other members of the thavazhi in favor of the\\ only other member and not an assignment of undivided shares.  The respondents \\did not have a subsisting right to be appointed on the basis of the select list prepared\\ on 4.4.87 as its life had expired. The Appointing Authority was justified in filling \\up the\end{tabular}} \\ \hline

  Case Text with Statutes \& Precedents &
  {\begin{tabular}[c]{@{}l@{}}Decision: Rejected \\Explanation:
  \\The appeal is accepted because the transaction Exhibit-B-9 is not void in law as it\\ is a release of rights by all the other members of the thavazhi in favor of the only\\ other member and not an assignment of undivided shares. \\
  ........................................................\\
  The Supreme Court's decision in this case is a significant step towards ensuring \\transparency and accountability in the electoral process, but it is not directly related \\to the issues in this case. Therefore, the appeal is accepted.  The decision in S. G. \\Jaisinghani vs Union Of India And Ors\end{tabular}} \\ \hline
\end{tabular}%
}
\caption{We observe `reject' judgments in the response generated by LLaMa-3x8B with varying supporting information.}
\label{cpt-hallucination-example}
\end{table*}



\begin{table*}[t]
\centering
\begin{tabular}{|p{0.95\textwidth}|}
\hline
\textbf{CASE NO:} \\ \hline
 CIVIL APPEAL NO………………. OF 2014\\
(Arising out of SLP (C) No.22035 of 2013)\\ \hline
\textbf{APPELLANTS:} \\ \hline
Usha Bharti  \\ \hline
\textbf{RESPONDENT:} \\ \hline
State Of U.P. \& Ors \\ \hline
\textbf{DATE OF JUDGMENT:} \\ \hline
28/03/2014 \\ \hline
{ \textbf{BENCH:}} \\ \hline
{ Fakkir Mohamed Ibrahim Kalifulla} \\ \hline
\textbf{CASE TEXT:} \\ \hline

{ \begin{tabular}[c]{@{}l@{}}
... The earlier judgment of the High Court in the writ petition clearly merged with the judgment of\\the High Court dismissing the review petition. Therefore, it was necessary only, in the peculiar\\facts of this case, to challenge only the judgment of the High Court in the review petition. It.... \\ \\ 

...These Rules can be amended by the High Court or the Supreme Court but \textcolor{blue}{Section 114} can only \\be amended by the Parliament. He points out that \textcolor{blue}{Section 121 and 122}, which permits the High \\Court to make their own rules on theprocedure to be followed in the High Court as well as in...  \\ \\ 

...The principle of Ejusdem Generis should not be applied for interpreting these provisions. \\Learned senior counsel relied on Board of Cricket Control (supra). He relied on Paragraphs 89, \\90 and 91. learned senior counsel also relied on \textcolor{magenta}{S. Nagaraj \& Ors. Vs. State of Karnataka \& Anr}\\.[13] He submits finally that all these judgments show that justice is above all. Therefore, no...\\ \\

... We are unable to accept the submission of Mr. Bhushan that the provisions contained in Section \\28 of the Act cannot be sustained in the eyes of law as it fails to satisfy the twin test of reasonable \\classification and rational nexus with the object sought to be achieved. In support of this submission, \\Mr. Bhushan has relied on the judgment of this Court in \textcolor{magenta}{D.S. Nakara vs. Union of India}[16]. We...\end{tabular}} \\ \hline
\textbf{JUDGEMENT:} \\ \hline
{ \begin{tabular}[c]{@{}l@{}}.... When the order dated 19th February, 2013 was passed, the issue with regard to reservation was\\ also not canvassed. But now that the issue had been raised, we thought it appropriate to examine \\the issue to put an end to the litigation between the parties.
\\ \\
In view of the above, \textcolor{red}{the appeal is accordingly dismissed}.....\end{tabular}} \\ \hline
\end{tabular}%

\caption{Example of Indian Case Structure. Sections referenced are highlighted in blue, previous judgments cited are in magenta, and the final decision is indicated in red.}
\label{case-example}
\end{table*}

\begin{table*}[ht]
    \centering
    \begin{tabular}{|p{0.8\textwidth}|}
    \hline
{\bf Template 1 (prediction + explanation)}\\
\hline
   
{\bf prompt} = f``````Task: Your task is to evaluate whether the appeal should be accepted (1) or rejected (0) based on the case proceedings provided below..\\

{\bf Prediction}: You are a legal expert tasked with making a judgment about whether an appeal should be accepted or rejected based on the provided summary of the (case/facts) along with (Precedents/statutes/both) depending on the pipeline. Your task is to evaluate whether the appeal should be accepted (1) or rejected (0) based on the case proceedings provided below. \\

  {\bf case\_proceeding}: \# case\_proceeding example 1\\

  {\bf Prediction}: \# example 1 prediction \\

  {\bf Explanation}: \# example 1 explanation\\

  {\bf case\_proceeding}: \# case\_proceeding example 2\\

 {\bf  Prediction}: \# example 2 prediction\\

 {\bf Explanation}: \# example 2 explanation\\

{\bf Instructions}: L\#\#\# Now, evaluate the following case:
        
Case proceedings: {summarized\_text}
        
Provide your judgment by strictly following this format:

\#\#PREDICTION: [Insert your prediction here]\\
\#\#EXPLANATION: [Insert your reasoning here that led you to your prediction.]
           
Strictly do not include anything outside this format. Strictly follow the provided format. Do not generate placeholders like [Insert your prediction here]. Just provide the final judgment and explanation. Do not hallucinate/repeat the same sentence again and 
again''''''\\
\hline

    \end{tabular}
    \caption{Prompts for Judgment Prediction.}
    \label{tab:judgment_prediction_prompts_few}
\end{table*}

\begin{table*}[ht]
\centering
\begin{tabular}{|l|}
\hline
\textbf{Instructions}:\\
You are an expert in legal text evaluation. You will be given:\\

A document description that specifies the intended content of a generated legal explanation.\\
An actual legal explanation that serves as the reference. A generated legal explanation that\\ needs to be evaluated.
Your task is to assess how well\ the generated explanation aligns with\\ the given description while using the actual document as a reference for correctness.\\
\\

\textbf{Evaluation Criteria (Unified Score: 1-10)}\\
Your evaluation should be based on the following factors:\\

\textit{Factual Accuracy (50\%)} – Does the generated document correctly represent the key legal\\ facts, reasoning, and outcomes from the original document, as expected from the description?\\
\textit{Completeness \& Coverage (30\%)} – Does it include all crucial legal arguments, case details,\\ and necessary context that the description implies?\\
\textit{Clarity \& Coherence (20\%)} – Is the document well-structured, logically presented,\\ and legally sound?\\
\\
\textbf{Scoring Scale:}\\
1-3 → Highly inaccurate, major omissions or distortions, poorly structured.\\
4-6 → Somewhat accurate but incomplete, missing key legal reasoning or context.\\
7-9 → Mostly accurate, well-structured, with minor omissions or inconsistencies.\\
10 → Fully aligned with the description, factually accurate, complete, and coherent.\\
\\
\textbf{Input Format:}\\
Document Description:\\
\{\{doc\_des\}\}\\
\\
\textbf{Original Legal Document (Reference):}\\
\{\{Actual\_Document\}\}\\
\\
\textbf{Generated Legal Document (To Be Evaluated):}\\
\{\{Generated\_Document\}\}\\
\\
\textbf{Output Format:}\\
Strictly provide only a single integer score (1-10) as the response,\\with no explanations, comments, or additional text.\\
\hline
\end{tabular}
\caption{The prompt is utilized to obtain scores from the G-Eval automatic evaluation methodology. We employed the GPT-4o-mini model to evaluate the quality of the generated text based on the provided prompt/input description, alongside the actual document as a reference.}
\label{tab:G-Eval Prompt}
\end{table*}